\documentclass[letterpaper, 10 pt, conference]{ieeeconf}
\IEEEoverridecommandlockouts
\usepackage[utf8]{inputenc}
\usepackage[T1]{fontenc}
\usepackage[english]{babel}
\usepackage{booktabs}
\usepackage{tabularx}
\usepackage{multirow}
\usepackage{listings}
\usepackage{subcaption}
\usepackage{xcolor}
\usepackage{xspace}
\usepackage{amsmath}
\usepackage{amssymb}
\usepackage{amsthm}

\newtheoremstyle{main}
{0.5em}                                                
{0.5em}                                                
{\itshape}                                           
{0pt}                                                
{\scshape}                                           
{}                                                   
{2pt}                                                
{\thmname{#1}\thmnumber{ #2}\thmnote{ {\normalfont(#3)}}:} 

\theoremstyle{main}
\newtheorem{definition}{Def.}[]

\DeclareMathOperator{\FK}{FK}
\DeclareMathOperator{\IK}{IK}
\DeclareMathOperator{\Proj}{Proj}

\usepackage{graphicx}
\usepackage{booktabs}
\usepackage{siunitx}

\usepackage{adjustbox}

\usepackage{csquotes}
\usepackage{algorithm}
\usepackage{algpseudocodex}

\usepackage{etoolbox}
\usepackage[
	maxbibnames=99,
	maxcitenames=2,
	natbib=true,
	style=numeric-comp,
	backend=biber,
	sorting=none,
	giveninits=true,
	url=false,
	doi=false,
	eprint=false,
	isbn=false,
]{biblatex}

\definecolor{purduegold}{HTML}{C28E0E} 

\makeatletter
\let\NAT@parse\undefined
\makeatother
\usepackage[pdfa,colorlinks,bookmarksopen,bookmarksnumbered]{hyperref}

\usepackage[nameinlink,capitalise]{cleveref}
\crefformat{line}{#2Ln.~#1#3}
\Crefformat{line}{#2Ln.~#1#3}
\makeatletter
\newcommand{\alglabel}[1]{%
  \cref@old@label{#1}%
  \protected@write\@auxout{}{\string\newlabel{#1@cref}{{[line][\arabic{ALG@line}][]\theALG@line}{[1][\thepage][]\thepage}{}{}{}}}}
\makeatother

\crefname{figure}{Fig.}{Figs.}
\Crefname{figure}{Fig.}{Figs.}
\crefname{equation}{Eq.}{Eqs.}
\Crefname{equation}{Eq.}{Eqs.}
\crefname{section}{Sec.}{Secs.}
\Crefname{section}{Sec.}{Secs.}
\crefname{definition}{Def.}{Defs.}
\Crefname{definition}{Def.}{Defs.}
\crefname{algorithm}{Alg.}{Algs.}
\Crefname{algorithm}{Alg.}{Algs.}
\crefname{table}{Tbl.}{Tbls.}
\Crefname{table}{Tbl.}{Tbls.}
\crefname{assumption}{Asm.}{Asms.}
\Crefname{assumption}{Asm.}{Asms.}
\crefname{subassumption}{Asm.}{Asms.}
\Crefname{subassumption}{Asm.}{Asms.}
\Crefname{problem}{Problem}{Problems}
\crefname{problem}{Problem}{Problems}

\usepackage{flushend}

\definecolor{setgreen}{HTML}{66c2a5}
\definecolor{setorange}{HTML}{D18F00}
\definecolor{setblue}{HTML}{8da0cb}
\definecolor{setpurp}{HTML}{e78ac3}
\definecolor{setorangevariant}{HTML}{cc4400}
\definecolor{setgreenorange}{HTML}{A6B153}

\definecolor{rblue}{HTML}{4169E1}
\newcommand{\contrib}[1]{\textcolor{rblue}{#1}}

\newcommand{\monogram}[3]{{}^{#2}\!#1^{#3}}
\usepackage[suppress]{color-edits}
\definecolor{darkgreen}{rgb}{0.0, 0.5, 0.0}
\addauthor{si}{blue}
\addauthor{tc}{darkgreen}
\addauthor{zk}{red}
\addauthor{sitotc}{orange}
\addauthor{todoshru}{brown}

\title{\LARGE \bf
ReVAMP: Vector-Accelerated Motion Planning \\ for Kinematically-Constrained Systems via Reparameterization
}

\newif\ifanonymous
\anonymousfalse

\ifanonymous
  \author{Anonymous Author(s)}
\else
\author{
Shrutheesh R. Iyer, Thomas Cohn, and Zachary Kingston%
\thanks{SRI, ZK are with the Department of Computer Science, Purdue University, {\tt \{iyer270,  zkingston\}@purdue.edu}. 
TC is with the Department of Electrical Engineering and Computer Science, Massachusetts Institute of Technology, {\tt tcohn@mit.edu}. This work was supported in part by the National Science Foundation Graduate Research Fellowship Program under Grant No. 2141064. Any opinions, findings, and conclusions or recommendations expressed in this material are those of the author(s) and do not necessarily reflect the views of the National Science Foundation.}%
\\
}
\fi

\begin{document}
\maketitle



\begin{abstract}
Robots often must satisfy one or more constraints during motion planning for real-world tasks. When such constraints reduce the valid configuration space to a measure-zero subset, sampling based planning algorithms require modifications to draw feasible samples. For many common end-effector constraints, parameterizations built on inverse kinematics (IK) provide an alternate formulation where the constraints are satisfied by construction, allowing directly sampling the feasible set. Despite their elegant approach, parameterized planners have remained slower than vector-accelerated implementations of projection-based approaches, leaving their performance ceiling an open question. We explore a new axis of vectorization built upon reparameterizing the planning space through analytic IK. This approach addresses existing inefficiencies in vectorized projection-based planners and exposes new opportunities for parallelism within the planner. We show that the planner can synthesize plans in microseconds to milliseconds for high dimensional systems (up to 20 dimensions), with complex constraints, up to 10x faster than the current state-of-the-art. Furthermore, we demonstrate how such planning speeds open up avenues for restructuring sequential manipulation pipelines. 
\end{abstract}

\section{Introduction}

\begin{figure}[t]
    \centering
    \includegraphics[width=0.45\linewidth]{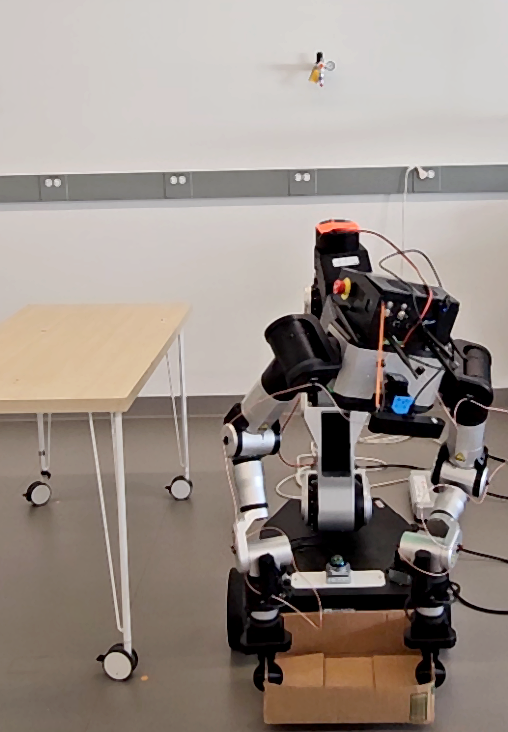}%
    \hfill
    \includegraphics[width=0.45\linewidth]{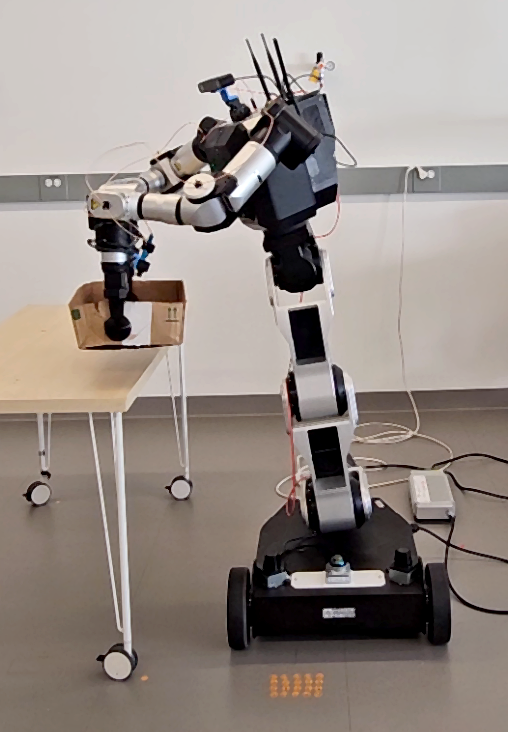}
    \caption{A 23-DoF bimanual mobile manipulator performing a whole-body pick-and-place task with motion plans generated by our motion planner.}
    \label{fig:ruby_box}
    \vspace{-1em}
\end{figure}

Several real-world robotics tasks require the robot to execute motions that respect constraints, leading to the well-studied problem of constrained motion planning (CMP). The most common class of these constraints realize themselves as end-effector constraints. For instance, \cref{fig:ruby_box} demonstrates a 23-DoF bimanual mobile manipulator carrying a box with both hands. Here, the relative transform between the two hands must remain constant throughout the transport motion, as determined by the grasp and the dimensions of the box. These constraints effectively shrink the space of valid configuration space into a zero-measure subset, often called a \emph{constraint manifold}, and computing plans through such a space is an expensive and non-trivial operation.


The ability to solve such constrained motion planning problems in real-time is still an open challenge.
Recent advances in vector-accelerated motion planning (VAMP)~\cite{vamp} and GPU parallelism~\cite{sundaralingam2023curobo1} have pushed planning times into the microsecond-to-millisecond range by parallelizing key subroutines in planning. Beyond reducing latency, these works can unlock new capabilities that require solving multiple motion planning problems, such as online replanning in dynamic environments and sequential manipulation tasks involving multiple planning queries~\cite{shen2024differentiable}. 
McVAMP~\cite{iyer2026vectorizing} extends the idea of vectorization to manifold constrained planning by employing vectorized projection operations of joint configurations onto the respective constraint manifolds. 
The central insight of VAMP and McVAMP is to decouple a robot's inherent properties (forward kinematics and differentials) from task-specific properties, such as the environment. These mappings are precompiled into hardware-specific routines, which are then vectorized using Single Instruction, Multiple Data (SIMD) CPU instructions, closing the real-time gap for CMP without sacrificing guarantees. 

Although the constraint manifold reduces the feasible space dimensionality, vectorized planners still operate in the full-dimensional ambient space, and rely on fast projections to accelerate  planning. We instead bring this vectorization paradigm to another class of CMP methods based on reparameterizing the planning space. These methods define a lower-dimensional planning space from the constraints, satisfying them by construction, reducing the dimensionality of the planning problem. For instance, a 14-DoF space, subjected to a bimanual constraint can be reduced into a 8-DoF planning space, avoiding sampling and projection in the ambient space. Parameterization using analytic inverse kinematics have been used to plan in task-space and then map back to joint configurations~\cite{cohn2024constrained}. 
The analytic reparameterization effectively eliminates the corresponding equality constraint from the motion planning problem, turning it into an unconstrained problem. However, these planners are often bottlenecked by the need to call IK hundreds to thousands of times for validation, which dramatically increases the planning time due to the latency of invoking the IK routine.

At the same time, this formulation lends itself more naturally to vectorized execution. Unlike iterative numerical projection whose computation time depends on the convergence, analytic IK evaluates a predetermined, non-iterative sequence of operations for each planning state. 
This is repeatable across planning states, making it well-suited to SIMD-style execution. 
Crucially, similar to forward kinematics, the analytic IK mapping is also a fundamental property intrinsic to the kinematics of a robot, and thus can be precompiled once for a given robot. Its implementation can be specialized and optimized for the underlying hardware. 

To this end, we explore a new axis of precompilation and subsequent vectorization for an accelerated planner that uses an analytic parameterization to plan in end-effector constrained spaces. Our proposed planner runs on a single core CPU and generates constraint-satisfying plans in real-time, ranging from microseconds for 7-DoF systems to milliseconds for a 20-DoF bimanual mobile manipulator. Planning in end-effector space further enables early collision detection and predictable per-iteration computation. We finally demonstrate how this increased throughput can enable new possibilities for sequential manipulation pipelines for a humanoid pick-and-place task. 


\section{Related Work}
Constraint motion planning is a well-studied problem due to its applicability to the real world. Several approaches have been proposed for CMP, which can broadly be classified into optimization-based, sampling-based, and learning-based methods. Optimization based methods~\cite{dragan2011manipulation, schulman2014motion,sundaralingam2023curobo1} typically treat the constraint as a soft-penalty term in their planning objective. But they can only satisfy the constraints weakly and can get stuck in local minima.
Sampling-based planning methods (SBMP) must leverage specialized routines to draw samples from the measure-zero constraint manifold~\cite{kingston2018sampling}, such as projection~\cite{berenson2009manipulation} or numerical continuation~\cite{jaillet2013atlasrrt, kim2016tangent}, which can all be abstracted away in standard SBMP methods~\cite{kingston2019exploring}.
However, sampling from constraint manifolds is often an expensive operation, they may satisfy constraints only up to $\epsilon$-tolerances, which may not be sufficient for sensitive systems. The distorted paths are also not straightforward for a controller to execute. 
Learning based methods attempt to learn a constraint, so that sampling and traversing the manifold can be done directly in the latent space of a neural network~\cite{qureshi2020neural, qureshi2021constrained, ni2024physics}.
But these methods require time-consuming retraining for new constraints, for which data collection may not be easy.

Yet another line of research attempts to directly sample the constraint manifold, using numerical or analytical IK to recover joint configurations~\cite{cortes2005sampling, han2001kinematics, wang2019inverse, lee2014prot}. Cohn et al.~\cite{cohn2024constrained} extend this idea by constructing parameterizations tailored to bimanual constraints and using the resulting IK mappings to reduce the dimensionality of the search space. Most of these methods leverage analytic IK routines to resolve parameterized configurations back into joint-space configurations for collision and validity checking. Tools such as IKFast~\cite{ikfast}, IK-Geo~\cite{elias2025ik} and SSIK~\cite{ssik} can be used to automatically generate analytic IK solutions for a wide range of arms. 
Yet, these methods are often bottlenecked by the IK computation at each step of planning to perform validity checking, which slows down the planner. Our work is strongly inspired by this class of methods, and attempts to specifically address the computational cost of IK during planning.

To improve the throughput of motion planning, works have focused extensively on using parallelism~\cite{amato1999probabilistic}, targeting different pieces of the pipeline. Some methods perform multiple SBMP iterations in parallel, while others run multiple searches in parallel~\cite{cforest}. cuRobo~\cite{sundaralingam2023curobo1} uses GPU parallelism to perform multi-seed optimization-based motion planning, treating constraints and collisions as penalty terms to the objective. VAMP~\cite{vamp} on the other hand takes a fine-grained approach and uses CPU SIMD operations to parallelize collision-checks for edge validation within a planning iteration, which has shown orders of magnitude performance boost. McVAMP~\cite{iyer2026vectorizing} similarly parallelizes the projection operations inside a planning iteration onto a manifold for constraint motion planning, which also demonstrates an equivalent speedup. However, it is still bottlenecked by the iterative projection procedure, which could arbitrarily stall the pipeline. Empirically, for a 7-DoF arm with a plane constraint, it spends 16\% of its projection time waiting for the last sample to converge.
We follow a similar approach, and attempt to use fine-grained parallelism to resolve the IK bottleneck, by performing analytic IK in parallel, which is more suited to vectorization.

\section{Preliminaries}
\label{sec:prelim}

Let $C \subset \mathbb{R}^d$ be the configuration space of a robot with $d$ degrees of freedom, where a configuration $q \in C$ is typically the vector of actuated joint angles.
In the presence of environment obstacles, $C_\text{free} \subset C$ denotes the subset of configurations that are not in collision.
A robot typically interacts with the environment through its end-effector(s), EE(s).
The forward kinematics map $\FK : C \to SE(3)$ gives the EE pose of a configuration and defines the task space reachable by the end-effector,
\begin{equation*}
  T = \FK(C) = \{\, \FK(q) \mid q \in C \,\} \subset SE(3).
\end{equation*}

In constrained motion planning (CMP), the path must additionally satisfy task constraints, typically expressed on $C$ by an implicit constraint function $F : C \to \mathbb{R}^k$, where $k$ is the dimensionality of the constraint.
This defines a lower-dimensional constraint manifold and its collision-free subset,
\begin{equation*}
  M = \{\, q \in C \mid F(q) = 0 \,\} \subset C, \quad
  M_\text{free} = M \cap C_\text{free}.
\end{equation*}

\begin{definition}[Constrained Motion Planning]\label{def:cmp}
Given $q_\text{init}, q_\text{goal} \in M_\text{free}$, find a continuous path $\sigma : [0,1] \to C$ such that
\begin{equation*}
  \sigma(0) = q_\text{init}, \quad
  \sigma(1) = q_\text{goal}, \quad
  \sigma(t) \in M_\text{free}, \forall t \in [0,1].
\end{equation*}
\end{definition}

Since $M$ has measure-zero in $C$, instead of uniform sampling, sampling-based CMP must generate samples on $M$ by other means~\cite{kingston2018sampling}.
Projection-based CMP~\cite{berenson2009manipulation} maps a sample $q \in C$ onto $M$ with an iterative operator $\Proj_M(q)$ that drives $F(q)$ to zero using Newton updates on $\partial F / \partial q$.

A widely used class of constraints restricts the EE pose to a subset $T_c \subset T$, itself defined implicitly by a function $g : SE(3) \to \mathbb{R}^k$,
\begin{equation*}
  T_c = \{\, X \in SE(3) \mid g(X) = 0 \,\}.
\end{equation*}
This induces the configuration-space constraint $F(q) := g(\FK(q))$, so that $M = \{\, q \in C \mid \FK(q) \in T_c \,\}$. This encompasses constraints ranging from keeping a glass level to fixing both feet of a humanoid to the ground.

Inverse kinematics is the preimage of a pose under $\FK$,
\begin{equation*}
  \IK : SE(3) \rightrightarrows C, \qquad
  \IK(X) = \{\, q \in C \mid \FK(q) = X \,\}.
\end{equation*}
Unlike $\FK$, $\IK$ is set-valued, as a single pose admits multiple solutions. Specifying a discrete branch $b$ and a continuous self-motion parameter $\psi \in \Psi$ gives a single-valued map $q = \IK_b(X,\psi)$
where $b$ selects a self-motion manifold (SMM) and $\psi$ parameterizes it~\cite{burdick1989inverse}.
This leads to an interpretation of $\IK$ as a nonlinear change of coordinates on $C$. 
Many EE constraints admit simple descriptions of $T_c$ (e.g., bimanual carrying is an affine constraint between the two EEs), so this change of coordinates yields a chart $M$.

When the end-effector constraint can be expressed functionally, the planning problem can be reparameterized~\cite{cohn2024constrained} to operate in a parameter space $\mathbb{P}$ instead of $C$.

\begin{definition}[Reparameterization]\label{def:reparam}
A reparameterization of $M$ is a pair $(\mathbb{P}, \Phi)$ of a parameter space $\mathbb{P}$ and a map $\Phi : \mathbb{P} \to C$, continuous where defined, such that $\Phi(\mathbb{P}) \subseteq M$.
\end{definition}

Every parameter thus satisfies the constraint by construction.
In our case, $\Phi$ is built from analytic $\IK$, e.g., for a redundant arm constrained to a plane with fixed orientation, $\mathbb{P} = \mathbb{R}^2 \times \Psi$ and $\Phi(p) = \IK_b(X(p), \psi)$, where $X(p) \in T_c$ is the pose determined by $p$.
Where $\IK_b$ has no solution (unreachable pose or joint limits), $\Phi$ is undefined and $p$ is invalid---we treat this similar to collision.
Conversely, a configuration $q \in M$ is mapped to $\mathbb{P}$ by $\Phi^{-1}(q)$, computed from $\FK(q)$ together with the branch and self-motion parameters recovered from $q$; this is how $q_\text{init}, q_\text{goal}$ are lifted to $p_\text{init}, p_\text{goal}$.
Sampling and interpolation in $\mathbb{P}$ are closed-form, and, unlike $M_\text{free} \subset C$, the collision-free parameter set
\begin{equation}\label{eqn:pfree}
  P_\text{free} = \{\, p \in \mathbb{P} \mid \Phi(p) \in M_\text{free} \,\}
\end{equation}
has positive measure in $\mathbb{P}$, so uniform sampling in $\mathbb{P}$ yields constraint-satisfying samples directly.

\begin{definition}[Reparameterized Motion Planning]\label{def:rmp}
Given a reparameterization $(\mathbb{P}, \Phi)$ and $p_\text{init}, p_\text{goal} \in P_\text{free}$ with $\Phi(p_\text{init}) = q_\text{init}$ and $\Phi(p_\text{goal}) = q_\text{goal}$, find a continuous path $\tau : [0,1] \to \mathbb{P}$ such that
\begin{equation*}
  \tau(0) = p_\text{init}, \quad
  \tau(1) = p_\text{goal}, \quad
  \tau(t) \in P_\text{free} \; \forall t \in [0,1].
\end{equation*}
\end{definition}

A solution $\tau$ to \cref{def:rmp} yields a solution $\sigma := \Phi \circ \tau$ to \cref{def:cmp}, since $\sigma$ is a composition of continuous maps.
The converse holds only within a single chart: a CMP solution $\sigma$ lifts to a path in $\mathbb{P}$ when $\Phi$ is a homeomorphism onto its image along $\sigma$, i.e., when $\sigma$ stays on one SMM branch $b$ away from kinematic singularities.
Solutions that cross branches are not captured by a single $(\mathbb{P}, \Phi)$; the parameterizations used for each robot are discussed in \cref{sec:expts}.

A noteworthy benefit of planning in the reparameterized space is that often, start-goal planning queries are easier to specify in the EE/parameterized space, rather than joint configurations on the constrained manifold space.

\section{Methodology}
In this work, we propose a new compilation axis for a robot: the analytic inverse kinematics routine, and the subsequent reparameterization that is constructed from it. Starting from the analytic solver, we transform the generated routines into branchless and vectorizable units that map parameterized configurations into joint configurations in parallel. This allows multiple configurations in the reparameterized space to be resolved simultaneously using SIMD parallelism. Finally, we integrate it into the sampling based planner, to accelerate the edge-validation of the planner.

\subsection{Analytic IK Compilation to generate the parameterization}
\label{sec:compileik}
The primary goal of reparameterization is to define a new planning space $\mathbb{P}$, along with mappings between it and the robot's joint configuration that enable planning operations. We generate this mapping using standard analytic IK tools such as IKFast~\cite{ikfast}. However, IKFast generates instructions with conditional statements that prevent trivial parallelization. Fortunately, these conditionals follow a specific structure that can be eliminated; consequently, we carefully modify in two stages to produce a branchless instruction set that is vectorizable. 
\begin{enumerate}
    \item At the highest level, it is split into $k$ templated initializations, one for each SMM branch defined by the sign for shoulder, wrist and elbow, so that no branch selection happens at runtime. According to ~\cite{burdick1989inverse}, there are at most 16 distinct branches where solutions lie for arms operating in $SE(3)$.
    \item For a fixed SMM branch, data-dependent logic such as guards for inverse trigonometric operations, angle-wraps, and wrist kinematic singularities are replaced with safe branchless versions, by relaxing them into a saturation loss instead of hard clamps. An IK solution is valid if the losses are all negative, and if the returned configuration lies within joint limits. This is analogous to the reachability probing functions in \cite[\S 4]{cohn2026framework}.
\end{enumerate}

These inverse kinematics are then used to generate the parameterized space, depending on the constraint-class. For a 7-DoF arm, it consists of the end-effector $SE(3)$ pose along with $\psi$. For a bimanual 14-DoF with a fixed relative transform constraint between the end-effectors, a single arm $SE(3)$ pose suffices. These parameterized spaces for each robot and constraint are discussed in~\cref{sec:expts}. 

The next step is to compile these parameterizations for planning. Similar to VAMP, we expand the tracing compiler built on top of Pinocchio~\cite{carpentier-sii19} and CppAD~\cite{cppad} to generate a branchless loop-free compiled instruction set for the IK and its subsequent parameterization.
Within the analytic IK function, inverse trigonometric operations are approximated using polynomials \tccomment{would we cite something here? I would appreciate as a reader}\sicomment{its from cephes which is ancient and industry standard}, and a smooth hinge loss is applied for reachability probes. We additionally compile the distance and interpolation operations on this new space. For pose components, we use spherical linear interpolation and a smooth $SO(3)$ geodesic, with their Euclidian counterparts for translation. For parameterizations that combine joints with IK~\cref{expts:ruby}, the interpolations and distance are split into linear interpolation and Euclidian distance for the joints, and $SE(3)$ operations for the pose.

A noteworthy benefit of compiling the IK function is its modularity: we can parameterize many constraint manifolds in terms of a single IK implementation, and therefore plan paths on them without recompliation.



\subsection{Vectorizing the trace compiled operations}

With this approach, the compiled instruction set now takes a parameterized space input $p \in \mathbb{P}$ and directly generates a joint configuration $q \in C$, along with validity, given by $\Phi : \mathbb{P} \to C$. In the sampling based planner, the sampled ``configurations", tree/graph structure, and motion validation, all exist within this new parameterized space. Since the space is constructed out of the task-enforced constraint by structure, all valid states in it naturally satisfy the constraint without extra operations. However, satisfying the constraint alone is not enough: conventional collision and joint-limit checking are still needed to ensure configuration validity.

Similar to VAMP, we vectorize the computation of IK using SIMD for its minimal overhead. Eliminating conditional logic enables us to repack the instruction set's input from an array-of-structures (AoS) into structure-of-arrays (SoA). The same IK routine can then process multiple parameterized space inputs simultaneously to obtain their corresponding joint configurations. The degree of parallelism is determined by the specific architecture the planner runs on; AVX for instance holds eight 32-bit floating points per register, letting us resolve 8 inputs in parallel.


\subsection{Accelerated Motion Planning using the parameterization}

\begin{figure} 
    \includegraphics[width=1.0\linewidth]{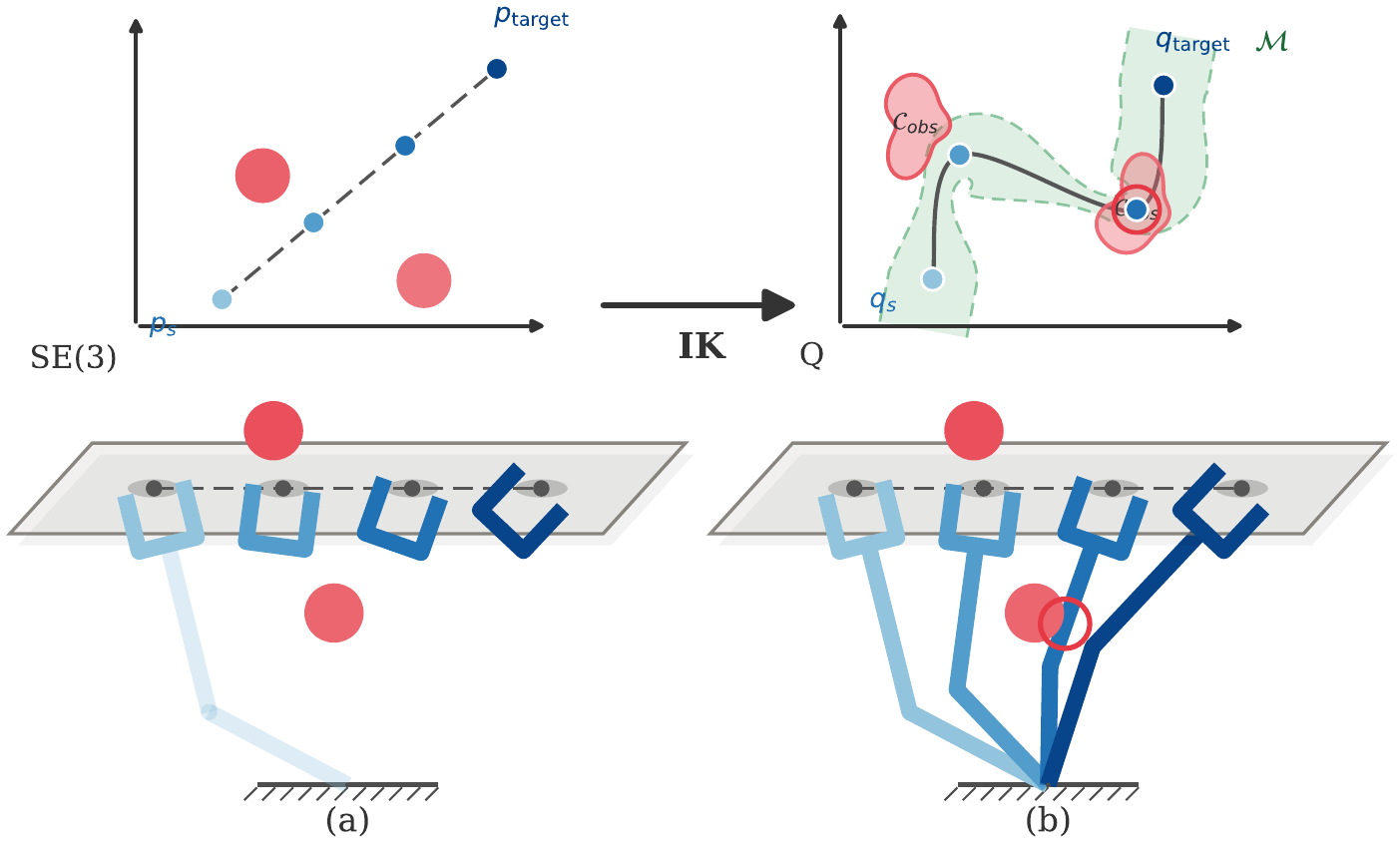}
    \vspace{-1em}
    \caption{Vectorized Edge-validation in ReVAMP. Given $p_s$ and $p_\text{target}$ in parameterized space, they are interpolated, that are then validated in parallel. There are two stages here (a) Early-collision checking pre-filter by eliminating edges where the interpolated P-space points are in collision with the environment. (b) If EEFs are collision free, then IK is resolved in parallel and a full C-space collision check is performed. \textit{$n=4$ (unit of parallelization) here for illustrative purposes}}
\label{fig:methodology}
\end{figure}

 \begin{algorithm}[!htbp]
\caption{\contrib{ParameterizedExtend}}
\vspace{0.2em}
\label{alg:paramextend}
\footnotesize
\begin{algorithmic}[1]
\Require $\mathcal{T}, p_s, p_{\text{target}}, \Phi, \mathcal{E}, r, \rho$
\Procedure{ParameterizedExtend}{}
\State $\text{dist} \gets \Call{Distance}{p_s, p_{\text{target}}}$
\State $\text{reach} \gets \text{dist} < r$
\If{reach}
    \State $p_{\text{next}} \gets p_{\text{target}}$
\Else
    \State \contrib{$p_{\text{next}} \gets \Call{Interpolate}{p_s, p_{\text{target}}, r / \text{dist}}$} \Comment{\contrib{steer in $\mathbb{P}$}} \alglabel{algline:pe:steer}
\EndIf
\If{$\neg \Call{Accept}{p_{\text{next}}}$}
    \State \Return \textsc{Rejected}
\EndIf
\For{$i = 1$ \textbf{to} $\lceil \text{dist} / (n \rho) \rceil$} \Comment{$n$-wide SIMD blocks} \alglabel{algline:pe:loop}
    \State $p^{(1:n)} \gets \Call{InterpolateBlock}{p_s, p_{\text{next}}, i}$ \alglabel{algline:pe:interpolateblock}
    \If{\contrib{$\neg \Call{EefCollisionFree}{p^{(1:n)}, \mathcal{E}}$}} \alglabel{algline:pe:eefprefilter}
        \State \contrib{\Return \textsc{Trapped}} \Comment{\contrib{IK-free collision prefilter}}
    \EndIf
    \State \contrib{$(\text{valid}, q^{(1:n)}) \gets \Phi\big(p^{(1:n)}\big)$} \Comment{\contrib{batched IK}} \alglabel{algline:pe:ik}
    \If{$\neg \text{valid} \lor \neg \Call{CC}{\mathcal{E}, q^{(1:n)}}$} \alglabel{algline:pe:fkcc}
        \State \Return \textsc{Trapped}
    \EndIf
    \If{\contrib{$\neg \Call{JointContinuous}{q^{(1:n)}, \delta_{\max}}$}} \alglabel{algline:pe:jointdelta}
        \State \Return \textsc{Trapped} \Comment{\contrib{interlane $\ell_\infty$ jump check}}
    \EndIf
\EndFor
\State $\Call{AddVertex}{\mathcal{T}, p_{\text{next}}}$
\State $\Call{AddEdge}{\mathcal{T}, p_s, p_{\text{next}}}$
\If{reach}
    \State \Return \textsc{Reached}
\Else
    \State \Return \textsc{Advanced}
\EndIf
\EndProcedure
\end{algorithmic}
\end{algorithm}

An SBMP typically spends a significant amount of its time validating edges connecting two states, to which we bring the vectorization for acceleration, as detailed in~\cref{alg:paramextend}
In the context of the parameterized planner, this requires sampling discrete points along the two states $p_s$ to $p_\text{target}$ up to a desired resolution~(\cref{algline:pe:interpolateblock}), transforming them back into configuration $q$ space to verify that (1) $q$ is valid and (2) $q$ is collision-free. First, we interpolate in the $\mathbb{P}$-space using closed-form interpolation operations that the parameterization provides between the start node $p_s$ and the target $p_\text{target}$~(\cref{algline:pe:interpolateblock}). For a motion/edge to be valid, every point along the edge needs to be valid.


To achieve this, we first use the vectorized IK and resolve operations across the different interpolated points to compute the joint configurations $q$~(\cref{algline:pe:ik}). Joint limits and reachability (using the saturation losses) can be checked in parallel from the IK without requiring forward kinematics. Then we use VAMP's vectorized broad-then-narrow-phase collision checking~(\cref{algline:pe:fkcc}) in parallel for the second condition. Since edge-validation has an all-or-nothing requirement, if any of the above checks fail even for a single point, the entire edge can be invalidated and exited early.


\textbf{Other tricks for efficiency:}
The end-effector is often the most-collision-prone link since it sweeps the largest space. Consequently, we first check the end-effector for collision at the sampled pose before IK~(\cref{algline:pe:eefprefilter}).
This prefilter is possible because our parameterized space directly carries the end-effector pose, unlike joint-space samples, which requires forward-kinematics. This eliminates a significant number of IK calls and further denser collision checking in cluttered environments. Finally, to ensure joint-continuity for smoothness, we reject an edge if any two consecutively resolved configurations differ by more than 0.15 rad in any joint~(\cref{algline:pe:jointdelta}). Besides helping to avoid singularities, this improves reliability as the planner is forced to admit edges only to a bounded distortion subset of the manifold.

\newsavebox{\mazetablebox}
\sbox{\mazetablebox}{%
  \begin{minipage}[b]{0.63\textwidth}
    \centering
    \scriptsize

    \label{tab:sim_maze_results}

    \begin{tabular}{lccc}
        \toprule
Metric & \textbf{\textcolor[HTML]{66C2A5}{McVAMP}} & \textbf{\textcolor[HTML]{FC8D62}{ReVAMP w/o EECC}} & \textbf{\textcolor[HTML]{8DA0CB}{ReVAMP}} \\
\midrule
Success (\%) & 88.5 & 89.0 & \textbf{89.5} \\
Iterations & 9442 (14644 $\pm$ 17872) & \textbf{5183 (11503 $\pm$ 16680)} & 5283 (11365 $\pm$ 17355) \\
Resolve (ms) & -- & 0.01 (0.01 $\pm$ 0.02) & \textbf{0.01 (0.01 $\pm$ 0.01)} \\
Planning (ms) & 46.19 (71.78 $\pm$ 92.21) & 5.52 (9.91 $\pm$ 11.91) & \textbf{4.38 (7.33 $\pm$ 8.72)} \\
Shortcut (ms) & 12.14 (16.75 $\pm$ 13.87) & 0.60 (0.69 $\pm$ 0.43) & \textbf{0.42 (0.48 $\pm$ 0.30)} \\
Total (ms) & 59.80 (88.53 $\pm$ 97.87) & 6.14 (10.60 $\pm$ 12.07) & \textbf{4.76 (7.81 $\pm$ 8.83)} \\
Failure (ms) & 473.22 (463.89 $\pm$ 68.89) & 75.01 (78.45 $\pm$ 19.56) & \textbf{57.31 (60.15 $\pm$ 18.17)} \\
Config dist. (rad) & \textbf{10.81 (10.75 $\pm$ 2.58)} & 11.57 (12.06 $\pm$ 4.14) & 11.49 (11.70 $\pm$ 3.74) \\
EEF dist. & 7.21 (7.30 $\pm$ 2.21) & 6.43 (5.71 $\pm$ 2.16) & \textbf{6.37 (5.47 $\pm$ 2.06)} \\
\bottomrule

\end{tabular}
    \captionof{table}{Maze Experiment Results: Median (Mean $\pm$ Std.) Planning Times and Path Lengths}
  \end{minipage}%
}

\begin{figure*}[!t]
\centering
\raisebox{\dp\mazetablebox}{\usebox{\mazetablebox}}%
\hfill
\begin{minipage}[b]{0.36\textwidth}
    \centering
    \includegraphics[width=\linewidth]{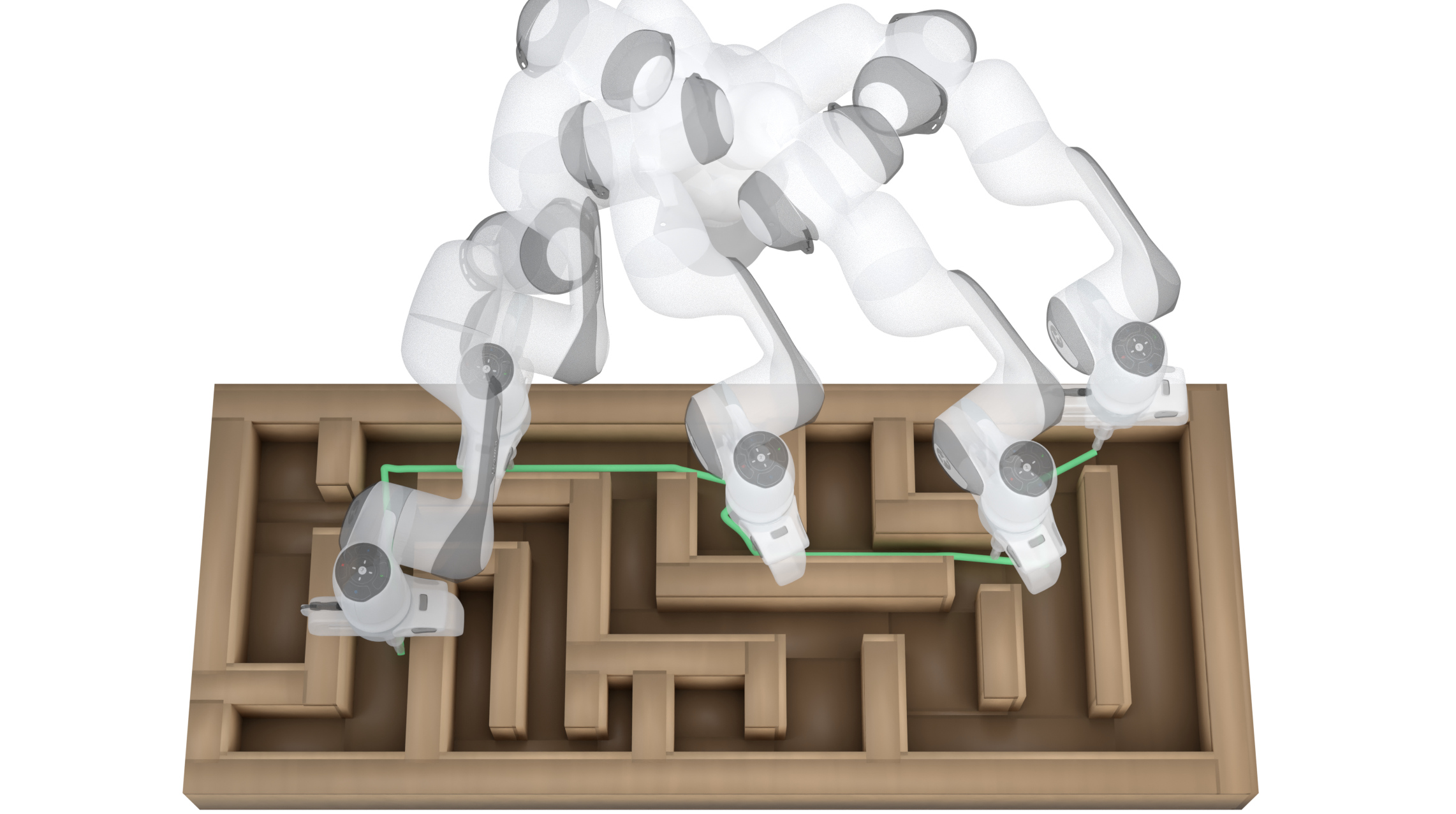}\par
    \captionof{figure}{Example of a solved trajectory in the maze.}
    \label{fig:maze_hero}
\end{minipage}
\end{figure*}

\section{Experimental Evaluation}
\label{sec:expts}
We evaluate our planner across different robots, with increasing DoF, and constraint complexity. All methods were evaluated on the same set of problems on a 5.4GHz Intel i7-13700K CPU with 64GB of RAM, unless specified otherwise. Each experiment highlights a distinct benefit of the planner and its potential implications. For the FR3 with a plane constraint, we show that planning on $SE(2)$ provides better scaling and strict constraint satisfaction. Through the bimanual 14-DoF Iiwa, we demonstrate the benefits of choosing the right parameterization, and the planner's high per-iteration throughput. Finally, on the RB-Y1 bimanual mobile manipulator, we show how the increased throughput can structurally change sequential planning pipelines. 



\subsection{Task Space Constraint for the 7-DoF FR3}



We first explore the task-space parameterization for the 7-DoF FR3 arm navigating a maze, with the end-effector constrained to the floor of the maze and its roll and pitch locked. We use the IK implementation provided by He. et al.~\cite{franka_ik2021}. It produces solutions in 4 different IK branches, with the angle of one joint used as the redundancy parameter $\psi$. Given a pose $X \in SE(3)$, $\psi \in [-\pi, \pi]$, and the desired branch $b$, it returns the configuration $q = \IK_b(X, \psi)$. The parameterized space is thus $\mathbb{P} = SE(3) \times \Psi$, which is 7-dimensional, with
\[
    \Phi : (X, \psi) \mapsto q.
\]

For the maze problem, $X$ is effectively reduced to sampling from $SE(2)$ since $z$, roll and pitch are fixed. To navigate from start to goal, it has to necessarily solve the maze. We generate 200 collision-free start-goal pairs belonging to the same SMM (such that they are sufficiently far apart). We compare it to the projection based McVAMP~\cite{iyer2026vectorizing} accelerated with the VAMP backend. We additionally ablate the early EEF collision checking (\cref{algline:pe:eefprefilter}), by disabling it in method "ReVAMP w/o EECC". Queries are terminated after 100k iterations without a solution. In addition to planning time and iterations, we record the time to resolve the $(x,y)$ goal to parameter space, and failure time.


\begin{figure}[h]
\centering
\begin{subfigure}[b]{0.75\linewidth}
    \centering
    \includegraphics[width=\linewidth]{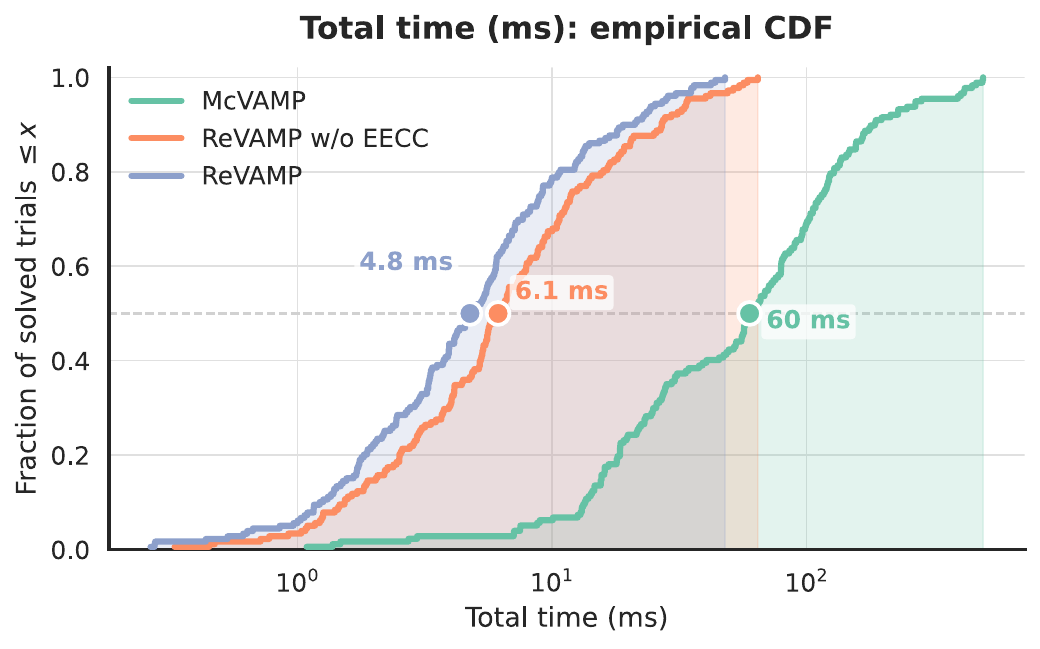}
    \caption{Total planning time (planning + shortcutting)}
    \vspace{0.5em}
    \label{tab:maze_time}
\end{subfigure}

\begin{subfigure}[b]{0.85\linewidth}
    \centering
    \includegraphics[width=\linewidth]{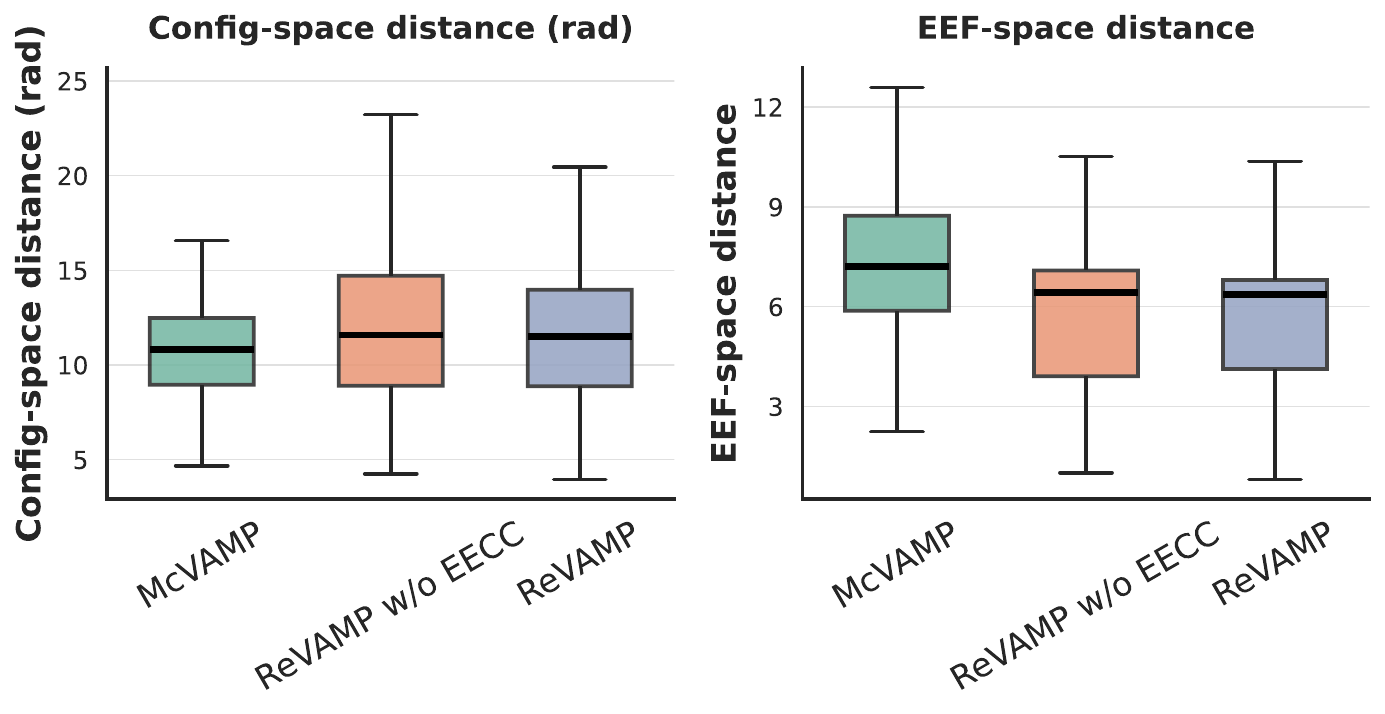}
    \caption{Config and EEF space distance}
    \label{tab:maze_distance}
\end{subfigure}
\caption{Maze solver results: (a) planning times to solve the maze (b) Task-space and configuration space distance}
\label{tab:maze_results}
\vspace{-1em}
\end{figure}

The results in ~\cref{tab:sim_maze_results} show that ReVAMP is better suited to the problem, with better planning times without sacrificing success. It scales better with harder problems, indicated by the long-tail planning times and a $10\times$ lower failure time. This strongly supports our hypothesis that the fixed non-iterative procedure allows failed queries to terminate much earlier. 
ReVAMP also requires only $(x,y)$ start-goal queries, while McVAMP requires constraint-satisfying configurations as start-goal states. \tccomment{Don't think you've discussed this yet?} The ablation shows the benefit of early collision checking: 91.6\% of edges are rejected before computing IK, reducing latency. Nonetheless, ReVAMP without this prefilter remains faster than McVAMP, indicating most of the gain comes from the reparameterization.


\begin{figure}
    \centering
    \includegraphics[width=0.95\linewidth]{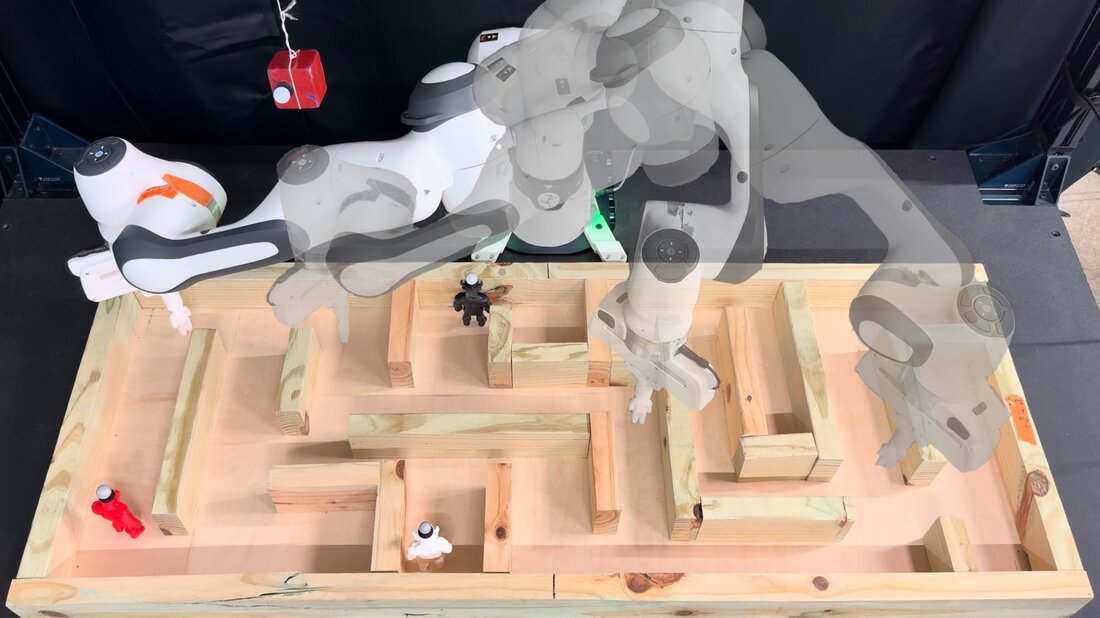}
    \caption{Real-world maze-setup, with obstacles that block the maze and the $C$-space. \sicomment{will try to do a ghosts in real-world, inspired by kaivalya} \tccomment{Might want to point claude at this code (overleaf comment) as an example of how to make a hardware swept volume figure.}}
    \label{fig:maze_hero_real}
    \vspace{-1em}
\end{figure}

\begin{table}[!t]
\centering
\scriptsize
\label{tab:real_maze_results}
\vspace{1em}
\begin{tabular}{lcc}
\toprule
Metric & McVAMP & ReVAMP \\
\midrule
Success (\%) & 91.89 & \textbf{95.65} \\
Planned z-error (mm) & 1.020 (1.227 $\pm$ 0.988) & \textbf{0.000 (0.000 $\pm$ 0.000)} \\
Executed z-error (mm) & 1.075 (1.307 $\pm$ 0.966) & \textbf{0.191 (0.221 $\pm$ 0.168)} \\
Plan. time (ms) & 146.19 (500.90 $\pm$ 666.05) & \textbf{11.95 (37.72 $\pm$ 67.07)} \\
Iterations & 10292 (30745 $\pm$ 39038) & \textbf{4570 (11930 $\pm$ 17295)} \\
\bottomrule
\end{tabular}
\caption{Real-Maze results. Values are median (mean $\pm$ std).}
\end{table}

We next evaluate the planners' validity in the real world, using the same maze setup~\cref{fig:maze_hero_real}, but with dynamic obstacles that block parts of the maze and the robot configuration space. The system must replan whenever the environment changes. We run both planners for approximately $60$s on an RTOS with a 2.5~GHz Intel i5-6500T CPU and 8~GB RAM, recording planning statistics and the executed joint trajectory at 30~Hz. We perform FK on the recorded states and measure the end-effector's deviation from the desired maze plane. \cref{tab:real_maze_results} clearly shows the benefits of our planner on a resource-constrained system. Beyond order-of-magnitude faster planning times, it importantly adheres to the task-constraint strictly. The planned trajectory remains on the desired plane, and the executed trajectory deviates by only $0.2$~mm on average (due to controller error).






\subsection{Bimanual Constraint for the 14-DoF Bimanual Iiwa}

We next test the system on a more complex constraint for a higher DoF system, the Bimanual 14-DoF Iiwa robot arms. Here, the relative pose between the end-effectors of the two arms is fixed (to the dimensions of the object it transports) throughout the motion. Given configurations $q_l$ and $q_r$ for the left and the right arms, and a fixed relative transform $\monogram{X}{l}{r}$ between the end-effectors, the constraint is $\FK(q_l)^{-1}\,\FK(q_r) = \monogram{X}{l}{r}$ . \tccomment{Might need to cite Russ' textbook for monogram notation (see comment for link).} We use the smooth IK solution proposed by Cohn et al. \cite{cohn2024constrained} for each arm and study two parameterizations built on it.

\paragraph{Leader-Follower Setup}
This is identical to the parameterization used by ~\cite{cohn2024constrained}: the leader arm samples in its configuration space $C_l$, while the follower arm's configuration is derived from the leader's end-effector pose, $q_r = \IK_b\!\left(\FK(q_l)\,{}^{l}X^{r}, \psi_r\right)$.
The parameterized space is therefore $\mathbb{P} = C_l \times \Psi$, where $q_l \in C_l$ is the leader's configuration and $\psi_r$ is the follower's free parameter. The 14-DoF configuration space is reduced to an \textbf{8-D} space, with the follower arm locked to a specified SMM branch. 
\[
    \Phi : (q_l, \psi_r) \mapsto (q_l, q_r).
\]

\paragraph{Dual-Follower Setup}
Given the object pose $\monogram{X}{W}{B}$, fixed grasp transforms ${}^{B}X^{l}$ and ${}^{B}X^{r}$, and $\psi_l$ and $\psi_r$ for each arm, we compute the desired end-effector pose of each arm, ${}^{W}X^{B}\,{}^{B}X^{l}$ and ${}^{W}X^{B}\,{}^{B}X^{r}$, and resolve both independently with $\IK_b$. The resulting space is $\mathbb{P} = SE(3) \times \Psi \times \Psi$, also an \textbf{8-D} parameterized space. Unlike the Leader-Follower setup, both arms are now locked to a single branch, which reduces the solution space. This is nevertheless desirable for tasks with additional constraints, such as keeping the box upright.
\[
    \Phi : ({}^{W}X^{B}, \psi_l, \psi_r) \mapsto (q_l, q_r).
\]



\begin{figure}[h]
\centering
\begin{subfigure}[b]{0.75\linewidth}
    \centering
    \includegraphics[width=\linewidth]{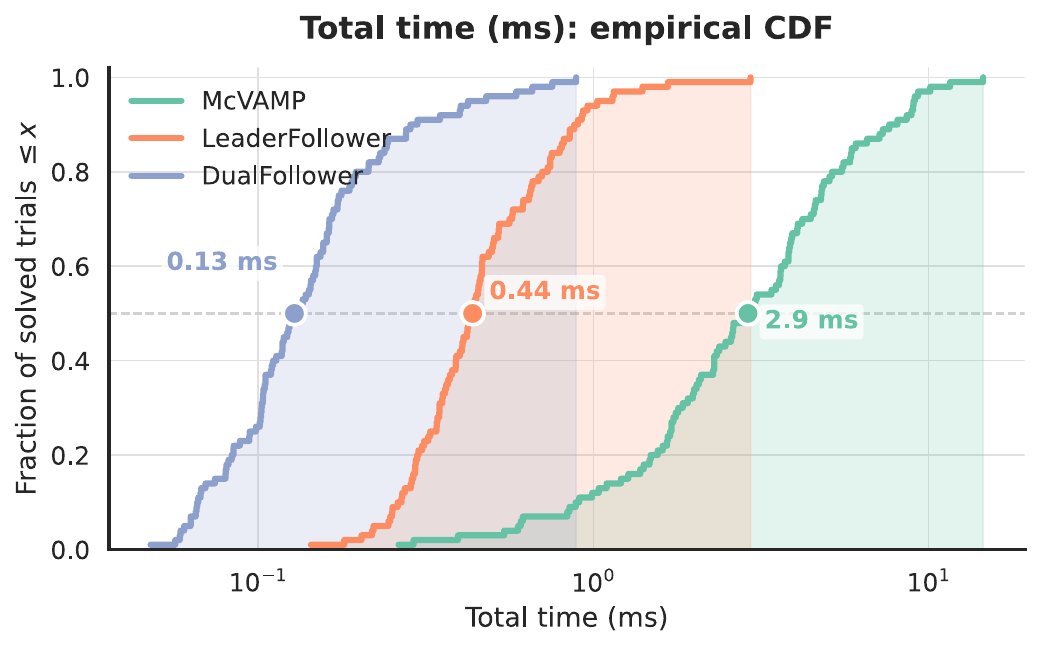}
    \caption{Total planning time (planning + shortcutting)}
    \vspace{0.5em}
    \label{fig:bimanual_time}
\end{subfigure}

\begin{subfigure}[b]{0.85\linewidth}
    \centering
    \includegraphics[width=\linewidth]{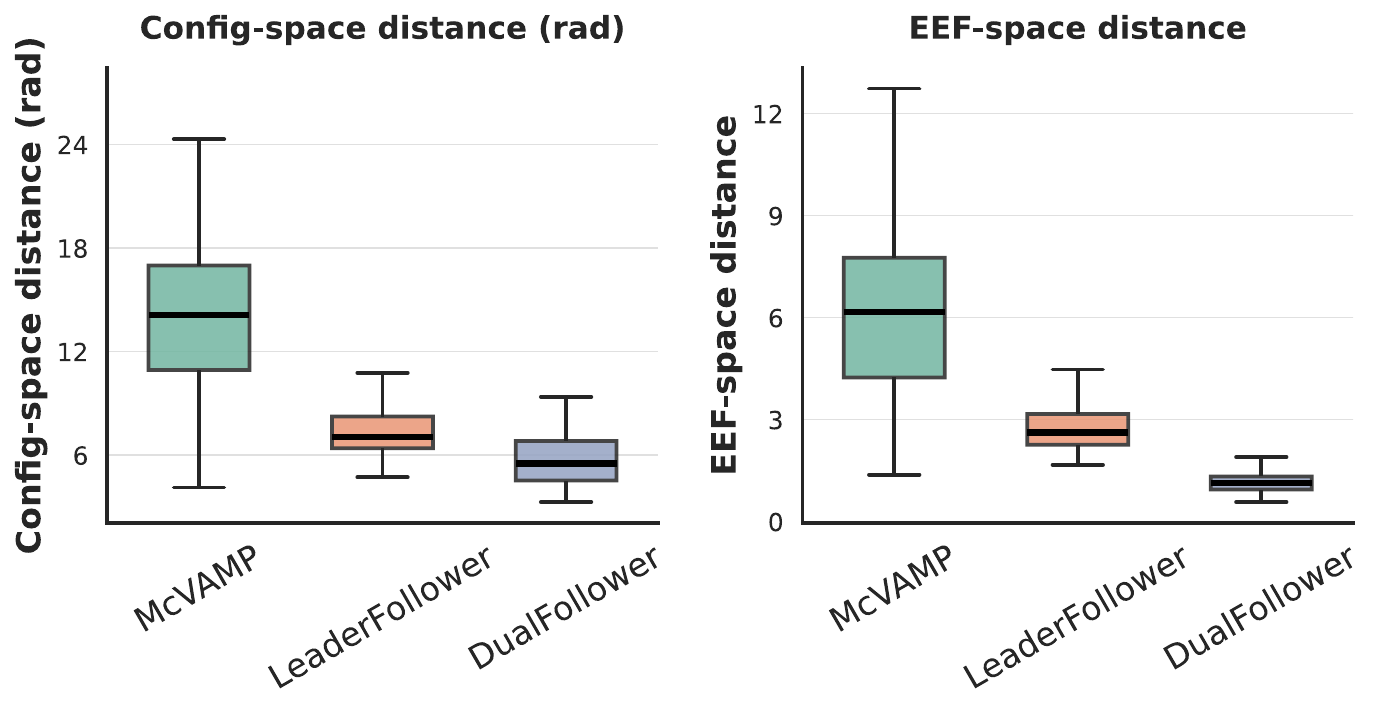}
    \caption{Config and EEF space distance}
    \label{fig:bimanual_distance}
\end{subfigure}
\caption{Bimanual transport results: (a) planning times (b) Task-space and configuration space distance \tccomment{Make sure to be consistent on the terminology for two followers (used ``Task-Space IK'' in previous section)}}

\label{fig:bimanual_results}
\vspace{-1em}
\end{figure}

\begin{table*}[!t]
\centering
\scriptsize
\vspace{0.2em}
\begin{tabular}{lcccccc}                                                                                                               \toprule
Method & Iterations & Planning (ms) & Shortcut (ms) & Total (ms) & Config dist. (rad) & EEF dist. \\       
\midrule
\textbf{\textcolor[HTML]{66C2A5}{McVAMP}} & 310 (437 $\pm$ 358) & 2.80 (3.56 $\pm$ 2.68) & 0.02 (0.07 $\pm$ 0.14) & 2.89 (3.63 $\pm$ 2.73) & 14.12 (14.03 $\pm$ 4.67) & 6.17 (6.22 $\pm$ 2.45) \\
\textbf{\textcolor[HTML]{FC8D62}{LeaderFollower}} & 1618 (2249 $\pm$ 2057) & 0.36 (0.45 $\pm$ 0.35) & 0.08 (0.09 $\pm$ 0.04) & 0.44 (0.53 $\pm$ 0.36) & 7.06 (7.30 $\pm$ 1.37) & 2.63 (2.77 $\pm$ 0.71) \\
\textbf{\textcolor[HTML]{8DA0CB}{DualFollower}} & \textbf{58 (75 $\pm$ 68)} & \textbf{0.11 (0.15 $\pm$ 0.13)} & \textbf{0.02 (0.02 $\pm$ 0.02)} & \textbf{0.13 (0.17 $\pm$ 0.14)} & \textbf{5.54 (5.68 $\pm$ 1.50)} & \textbf{1.14 (1.17 $\pm$ 0.32)} \\
\bottomrule
\end{tabular}
\caption{Median (mean $\pm$ std) planning cost and shortcut path length per method.}
\label{tab:bimanual_results_tab}
\end{table*}

We use the same problem set proposed by ~\cite{cohn2024constrained} between three fixed positions, with 200 trials, randomizing the start-goal pairs. ~\cref{tab:bimanual_results_tab} compares McVAMP with the Leader-Follower and Dual-Follower parameterizations. From the results in~\cref{tab:bimanual_results_tab}, it is clear that the parameterized planners achieve a significantly lower per-iteration planning time, particularly evident for the LeaderFollower. It is $6\times$ faster than McVAMP despite taking $5\times$ more iterations. This suggests that the parameterized planner scales much better with the dimensionality and the constraint complexity. Both planners also show lower planning times and trajectory distances. 
Interestingly, the Dual-Follower parameterization outperforms the Leader-Follower across all metrics. This is likely because all start-goal pairs lie in the same SMM for both arms, making the problem easier. Beyond vectorization, this experiment demonstrates that choosing the right planning space can significantly reduce the planning iterations, making the problem easier to solve in the new space.



\subsection{Pick and Place for RB-Y1}
\label{expts:ruby}
Our final experiment deploys our planner to generate whole-body motions for a full pick-and-place pipeline on the Rainbow Robotics RB-Y1 mobile manipulator. We use the parameterization based on the Inverse Function Theorem (IFT) work proposed by~\cite{cohn2026planning} . With a slight abuse of terminology, for the rest of the discussion, we refer to the entire pick-and-place pipeline method used by the work as IFT. It has two 7-DoF arms attached to a 6-DoF torso, and a wheeled base. Since we are evaluating a pure manipulation task, we disable the wheels, and fix the base at origin, which makes it a 20-DoF system. Again, the goal is a bimanual-constrained box-transport task, whose constraints are similar to the 14-DoF Iiwa arm. The robot must pick up a box from the floor with both hands and place it on a table next to the robot. It must also satisfy a stability constraint requiring the center of mass (CoM) to remain within the support polygon.


We use IKFast~\cite{ikfast} to generate the IK solution for each arm, and compile them as described in~\cref{sec:compileik}.
The arms each have 8 disconnected self-motion manifolds, with redundancy $\psi$ parameterized by the angle of the $3$rd joint.
We implement a Dual-Follower parameterization: the torso joint configuration, the held box pose, and free parameters $\psi_l$ and $\psi_r$ for each arm define a 14-DoF space,
\[
    \Phi : (q_\text{torso}, {}^{W}X^{B}, \psi_l, \psi_r) \mapsto (q_\text{torso}, q_l, q_r),
\]
where $B$ is the box frame.
As with IFT, we test 20 box poses in a 4-by-5 grid (3cm apart) on the floor in front of the robot. This experiment demonstrates the scaling of our method along multiple axes. The constrained problem has only 0.012\% of its parameterized space feasible under collision and stability constraints (an inequality constraint defined by support polytopes), as opposed to the 33.1\% of the C-space available for unconstrained planning.


We retain the complete planning pipeline from IFT~\cite{cohn2026planning}, which at a high level includes: (1) optimization IK for the pick, carry, and place poses, (2) unconstrained plan from start-to-pre-pick configuration, and place-to-home configuration, (3) constrained motions for pick to carry waypoint followed by carry to place and (4) trajectory optimization to improve the plan and TOPPRA for time parameterization.

We differ from IFT in two significant ways.
The first change is straightforward: the use of our vectorized planner to construct initial guesses for the constrained planning legs. Second, and more significantly, we change how pick and place poses are selected. IFT generates a collection of configurations and uses problem-specific heuristics to select candidates for motion planning, with fallbacks when these candidates are not plannable.
In contrast, we solve a constrained motion planning problem between \emph{every} triple of pick, carry, and place configurations\footnote{
    This only requires solving $2n^2$ planning problems, since the plans between pick and carry are independent of the plans between carry and place.
}, and then simply select the configurations yielding the shortest plan.
This is faster, simpler, and provides better motion plans, and is wholly enabled by the incredible speed of our vectorized planner.
Additionally, to evaluate the planner's benefits, we run 20 direct pick-to-place trials without the ``carry" waypoint used in IFT, simplifying the planning problem. \tccomment{Why ``retraction''? Maybe ``waypoint''?}
\begin{table}[!t]
  \centering
  \renewcommand{\arraystretch}{0.9}
  \scriptsize
  \label{tab:ruby_results}
  \begin{tabularx}{\columnwidth}{@{}>{\raggedright\arraybackslash}p{3.3cm}@{\hspace{2pt}}*{3}{>{\centering\arraybackslash}X}@{}}
  \toprule
  \textbf{Metric} & \shortstack{\textbf{IFT}\\\textbf{(baseline)}} & \shortstack{\textbf{Mod. IFT}\\\textbf{+ McVAMP}} & \shortstack{\textbf{Mod. IFT}\\\textbf{+ ReVAMP}} \\
  \midrule
  \multicolumn{4}{@{}l}{\textit{Pipeline Result}} \\
  Success Rate                 & 98\%~(39/40) & 95\%~(38/40) & \textbf{100\%~(40/40)} \\
  Time to Plan Median (s)      & 50.2 & 36.7 & \textbf{21.2} \\
  Time to Plan Mean (s)        & 70.2 & 41.8 & \textbf{25.0} \\
  Time to Plan Max (s)         & 218.2 & 84.6 & \textbf{75.5} \\
  Path Length (rad)            & 17.25 & \textbf{16.83} & 17.95 \\
  \midrule
  \multicolumn{4}{@{}l}{\textit{Aggregated Constrained Planning Statistics}} \\
  Total Planning Calls         & 142 & \textbf{1696} & \textbf{1696} \\
  Per-Call Success Rate        & 86\% & 52\% & 47\%  \\
  Succ.\ Call Time, mean (ms)           & 512.96 & \textbf{3.91} & 8.98 \\
  Succ.\ Call Time, max (ms)           & 5100.96 & \textbf{19.34} & 34.37 \\
  Unsucc.\ Call Time, mean (ms)           & 30003.51 & 246.40 & \textbf{8.89} \\
  Unsucc.\ Call Time, max(ms)           & 30022.4 & 1119.05 & \textbf{121.54} \\
  \midrule
  \multicolumn{4}{@{}l}{\textit{Per-Stage Results (s), mean / max}} \\
  Sampling-Based Planning      & 19.7 / 70.1 & 5.7 / 19.4 & \textbf{1.4 / 2.6} \\
  Optimization IK              & 16.8 / 33.2 & 7.2 / 27.1 & \textbf{4.5 / 15.8} \\
  Trajectory Opt.\ (Guess)     & \textbf{5.4 / 8.4} & 6.5 / 9.1 & 6.7 / 8.8 \\
  Trajectory Opt.\ (Solve)     & 16.3 / 180.2 & 17.5 / 42.2 & \textbf{7.7 / 55.5} \\
  TOPPRA                       & 2.6 / 3.5 & \textbf{1.1 / 1.5} & 1.4 / 1.5 \\
  \bottomrule
  \end{tabularx}
  \caption{
      RB-Y1 box pickup planning results: the original IFT pipeline~\cite{cohn2026planning} as a baseline, versus our modified pipeline instantiated with McVAMP and ReVAMP as the constrained-planning backend.
  }

\end{table}

We rerun the original IFT pipeline as the baseline and compare its results with our modified pipeline in~\cref{tab:ruby_results}, as well as the same pipeline using McVAMP as the constrained planner backend. Our framework produces shorter planning times with slightly worse path quality and solves the harder direct pick-to-place tasks more successfully. Although McVAMP has slightly lower atomic planning latency for successful plans, it takes longer to fail, which would be exacerbated by the large number of planning calls we make, the complete ReVAMP pipeline is faster because it produces trajectories that are easier to optimize.

More importantly, we show that these speedups are so great in magnitude that we can successfully restructure our pipelines to take advantage of the fact that satisficing planning has negligible runtime. The modified pipeline makes $10\times$ more motion-planning queries, yet the aggregated planning time is nearly $20\times$ lower. 
Constrained planning accounts for only 1\% of the per-point wall time, despite the increased planning queries. The pipeline as a whole became 2.5 times faster end-to-end on average.


\section{Conclusion and Future Work}

Recent works have shown that precompiling a robot's intrinsic mappings can significantly accelerate motion planning. In this work, we present a new class of such structures, built from analytic parameterizations of the planning space that satisfy end-effector constraints by construction. By exploiting known robot kinematic structures, we generate analytic IK routines ahead of time, and instantiate it for end-effector constrained planning by vectorizing it in a SIMD fashion. We show that for end-effector constraints, this formulation is particularly amenable to parallelism and strictly satisfies constraints, producing plans up to an order of magnitude faster. Through this, we explore other benefits that vectorizing the parameterized space can offer, such as planning in real-time with dynamic environments. We also show how an existing sequential planning pipeline can be modified to exploit this increased throughput.

Although our method can outperform its projection-based counterparts, it currently only applies when the entire plan lies on a single self-motion-manifold (SMM) branch, something that the projection-based planner is not restricted by. 
Surprisingly, for many real-world tasks a single IK branch covers enough of the space that this has little practical effect outside some long-tail cases. 
Nonetheless, we seek to explore being able to cross IK branches while planning, while still maintaining the benefits the planner provides.
In addition, modifying the generated IK solution is a tedious process, even though it is one-time and can be made easier with coding agents. 
We look to automate this generation by revisiting the analytic IK solvers themselves.

More broadly, we believe this new form of vectorization opens avenues across manipulation and planning. By precompiling differentials along with the IK, we can also explore optimal IK computation and grasp selection, replacing the need for heuristics. Finally, we see the largest applicability of our vectorized planner in task-and-motion-planning (TAMP) problems, where we can fundamentally revisit motion sampling through this new lens of cheap, constrained motion generation to bring completeness to TAMP.

\section*{Acknowledgment}
Claude code was used for editorial assistance and figure generation. All technical content, results, and final manuscript text were reviewed and verified by the authors.

\printbibliography{}
\end{document}